# Bayesian Approach to Rough Set

Tshilidzi Marwala and Bodie Crossingham

University of the Witwatersrand

Private Bag x3

Wits, 2050

South Africa

e-mail: t.marwala@ee.wits.ac.za

This paper proposes an approach to training rough set models using Bayesian framework trained using Markov Chain Monte Carlo (MCMC) method. The prior probabilities are constructed from the prior knowledge that good rough set models have fewer rules. Markov Chain Monte Carlo sampling is conducted through sampling in the rough set granule space and Metropolis algorithm is used as an acceptance criteria. The proposed method is tested to estimate the risk of HIV given demographic data. The results obtained shows that the proposed approach is able to achieve an average accuracy of 58% with the accuracy varying up to 66%. In addition the Bayesian rough set give the probabilities of the estimated HIV status as well as the linguistic rules describing how the demographic parameters drive the risk of HIV.

## Introduction

Rough set theory (RST) was introduced by Pawlak (1991) and is a mathematical tool which deals with vagueness and uncertainty. It is of fundamental importance to artificial intelligence (AI) and cognitive science and is highly applicable to the tasks of machine learning and decision analysis. Rough set are useful in the analysis of decisions in which there are inconsistencies. To cope with these inconsistencies, lower and upper approximations of decision classes are defined (Inuiguchib and Miyajima, 2006). Rough set theory is often contrasted to compete with fuzzy set theory (FST), but it in fact complements it. One of the advantages of RST is that it does not require a priori knowledge about the data set, and it is for this reason that statistical methods are not sufficient for determining the relationships that exists in complex cases such as between the demographic variables and their respective HIV status.

Greco et al. (2006) generalized the original idea of rough set and introduced variable precision rough set, which is based on the concept of relative and absolute rough membership. The Bayesian framework is a tool that can be used to extend this absolute to relative membership framework. Nashino et. al. (2006) proposed a rough set method to analyze human evaluation data with much ambiguity such as sensory and feeling data and handles totally ambiguous and probabilistic human evaluation data using a probabilistic approximation based on information gains of equivalent classes. Slezak and Ziarko (2005) proposed a rough set model, which is concerned primarily with algebraic properties of approximately defined sets and extended the basic rough set theory to incorporate probabilistic information.

This paper extends the rough set model to the probabilistic domain using Bayesian framework, Markov Chain Monte Carlo simulation and Metropolis algorithms. In order to achieve this, the rough set membership functions' granulizations are interpreted probabilistically. The proposed framework is then applied to modeling the relationships between demographic properties and the risk of HIV.

**Rough Set Theory (RST)**

Rough set theory deals with the approximation of sets that are difficult to describe with the available information (Orhn and Rowland, 2006). It deals predominantly with the classification of imprecise, uncertain or incomplete information. Some concepts that are fundamental to RST theory are given in the next few sections. The data is represented using an information table and an example for the HIV data set for the $i^{th}$ object is given in Table 1:

Table 1: Information Table of the HIV Data

|  | Race | Mothers' Age | Education | Gravidity | Parity | Fathers' Age | HIV Status |
|---|---|---|---|---|---|---|---|
| Obj$^{(1)}$ | 1 | 32 | 1 | 1 | 2 | 34 | 0 |
| Obj$^{(2)}$ | 2 | 27 | 13 | 2 | 1 | 28 | 1 |
| Obj$^{(3)}$ | 2 | 25 | 8 | 2 | 0 | 23 | 1 |
| . | . | . | . | . | . | . | . |
| Obj$^{(i)}$ | 3 | 27 | 4 | 3 | 1 | 22 | 0 |

In the information table, each row represents a new case (or object). Besides HIV status, each of the columns represents the respective case's variables (or condition attributes). The HIV Status is the outcome (also called the concept or decision attribute) of each object. The outcome contains either a 1 or 0, and this indicates whether the particular case is infected with HIV or not.

Once the information table is obtained, the data is discretised into partitions as mentioned earlier. An information system can be understood by a pair $\Lambda = (U, A)$, where U and A, are finite, non-empty sets called the universe, and the set of attributes, respectively (Deja and Peszek, 2003). For every attribute $a$ an element of A, we associate a set $V_a$, of its values, where $V_a$ is called the value set of $a$.

$$a: U \rightarrow V_a \tag{1}$$

Any subset *B* of *A* determines a binary relation *I(B)* on *U*, which is called an indiscernibility relation. The main concept of rough set theory is an indiscernibility relation (indiscernibility meaning indistinguishable from one another). Sets that are indiscernible are called elementary sets, and these are considered the building blocks of RST's knowledge of reality. A union of elementary sets is called a crisp set, while any other sets are referred to as rough or vague. More formally, for given information system $\Lambda$, then for any subset $B \subseteq A$, there is an associated equivalence relation *I(B)* called the $B - indiscernibility$ relation and is represented as shown as:

$$(x, y) \in I(B) \; iff \; a(x) = a(y) \tag{2}$$

RST offers a tool to deal with indiscernibility and the way in which it works is, for each concept/decision *X*, the greatest definable set containing *X* and the least definable set containing *X* are computed. These two sets are called the lower and upper approximation respectively. The sets of cases/objects with the same outcome variable are assembled together. This is done by looking at the "purity" of the particular objects attributes in relation to its outcome. In most cases it is not possible to define cases into crisp sets, in such instances lower and upper approximation sets are defined. The lower approximation is defined as the collection of cases whose equivalence classes are fully contained in the set of cases we want to approximate (Ohrn and Rowland, 2006). The lower approximation of set X is denoted $\underline{B}X$ and is mathematically represented as:

$$\underline{B}X = \{x \in U : B(x) \subseteq X\} \tag{3}$$

The upper approximation is defined as the collection of cases whose equivalence classes are at least partially contained in the set of cases we want to approximate. The upper approximation of set X is denoted $\overline{B}X$ and is mathematically represented as:

$$\overline{B}X = \{x \in U : B(x) \cap X = \emptyset\} \tag{4}$$

It is through these lower and upper approximations that any rough set is defined. Lower and upper approximations are defined differently in various literature, but it follows that a crisp set is only defined for $\overline{B}X = \underline{B}X$. It must be noted that for most cases in RST, reducts are generated to enable us to discard functionally redundant information (Pawlak, 1991) and in this paper the prior probability handles reducts.

*Rough Membership Function*

The rough membership function is described; $\eta_A^X : U \to [0, 1]$ that, when applied to object x, quantifies the degree of relative overlap between the set X and the indiscernibility set to which x belongs. This membership function is a measure of the plausibility of which an object x belongs to set X. This membership function is defined as:

$$\eta_A^X = \frac{|[X]_B \cap X|}{[X]_B} \tag{5}$$

where $[X]_b$ is an elementary set.

*Rough Set Accuracy*

The accuracy of rough set provides a measure of how closely the rough set is approximating the target set. It is defined as the ratio of the number of objects which can be positively placed in *X* to the number of objects that can be possibly placed in *X*. In other words it is defined as the number of cases in the lower approximation, divided by the number of cases in the upper approximation; $0 \leq \alpha_p(X) \leq 1$

$$\alpha_p(X) = \frac{|\underline{B}X|}{|\overline{B}X|} \tag{6}$$

*Rough set Formulation*

The process of modeling the rough set can be broken down into five stages. The first stage would be to select the data while the second stage involves pre-processing the data to ensure that it is ready for analysis. The second stage involves discretizing the data and removing unnecessary data (cleaning the data). If reducts were considered, the third stage would be to use the cleaned data to generate reducts. A reduct is the most concise way in which we can discern object classes (Witlox and Tindermans, 2004). In other words, a reduct is the minimal subset of attributes that enables the same classification of elements of the universe as the whole set of attributes (Pawlak, 1991). To cope with inconsistencies, lower and upper approximations of decision classes are defined (Ohrn, 2006; Deja and Peszek, 2003). Stage four is where the rules are extracted or generated. The rules are normally determined based on condition attributes values (Goh and Law, 2003). Once the rules are extracted, they can be presented in an *if CONDITION(S)-then DECISION* format (Leke, 2007). The final or fifth stage involves testing the newly created rules on a test set to estimate the prediction error of the rough set model. The equation representing the mapping between the inputs to the output using rough set can be written as

$$y = f(G, N, R) \tag{7}$$

where y is the output, $G$ is the granulization of the input space into high, low, medium etc, $N$ is the number of rules and $R$ is the rule. So for a given nature of granulization, the rough set model will be able to give the optimal number of rules and the accuracy of prediction. Therefore, in rough set modeling there is always a trade-off between the degree of granulization of the input space (which affects the nature and size of rules) and the prediction accuracy of the rough set model. Therefore, the estimation process for the level and nature of the granulization process will be solved using Bayesian framework, which is explained in the next section.

*Bayesian Training on Rough Set Model*

The Bayesian framework can be written as in (Marwala, 2007; Bishop, 2006):

$$P(M \mid D) = \frac{P(D \mid M) p(M)}{p(D)} \quad \text{where} \quad M = \begin{Bmatrix} G \\ N \\ R \end{Bmatrix} \tag{8}$$

Within the context of Bayesian rough set models, $G$ is granulization, $R$ = rough set rules, $N$ = number of rules, $D$ is the data which consist of input $x$ and output $y$ and $A$ = accuracy of rough set model prediction. The parameter $P(M \mid D)$ is the probability of the rough set model given the observed data, $P(D \mid M)$ is the probability of the data given the assumed rough set model and is also called the likelihood function, $P(M)$ is the prior probability of the rough set model and $P(D)$ is the probability of the data and is also called the evidence. The evidence can be treated as the normalization constant and therefore is ignored in this paper. The likelihood function may be estimated as follows:

$$P(D|M) = \frac{1}{z_1}\exp(-error) = \frac{1}{z_1}\exp\{A(N,R,G) - 1\} \quad (9)$$

Here $z_1$ is the normalization constant. The prior probability in this problem is linked to the concept of reducts, which was explained earlier and it is the prior knowledge that the best rough set models are the ones with the minimum numbers of rules ($N$). Therefore, the prior probability may be written as follows:

$$P(M) = \frac{1}{z_2}\exp\{-\lambda N\} \quad (10)$$

where $z_2$ is the normalization constant and $\lambda$ is a hyperparameter that scales the prior information to be in line with the magnitude of the likelihood function. The posterior probability of the model given the observed data is thus:

$$P(M|D) = \frac{1}{z}\exp\{A(N,R,G) - 1 - \lambda N\} \quad (11)$$

where $z$ is the normalization constant. Since the number of rules and the rules themselves given the data depend on the nature of the granulization of the input space, we shall sample in the granule space using a procedure called Markov Chain Monte Carlo simulation (Marwala, 2007; Bishop, 2006).

*Markov Monte Carlo Simulation*

The manner in which the probability distribution in equation 11 may be sampled is to randomly generate a succession of granule vectors and accepting or rejecting them based on how probable they are using Metropolis algorithm. This process requires a generation of large samples of granules for the input space, which in many cases is not computationally efficient. The MCMC creates a chain of granules and accepts or rejects them using Metropolis algorithm. The application of Bayesian approach and MCMC rough set, results in the probability distribution function of the granules, which in turn leads to the distribution of the rough set outputs. From these distribution functions the average prediction of the rough set model and the variance of that prediction can be calculated. The probability distributions of the rough set model represented by granules are mathematically described by equation 11. From equation 11 and by following the rules of probability theory, the distribution of the output parameter, *y*, is written as (Marwala, 2007):

$$p(y|x,D) = \int p(y|x,M)p(M|D)dM \tag{12}$$

Equation 12 depends on equation 11, and is difficult to solve analytically due to relatively high dimension of granule space. Thus the integral in equation 12 may be approximated as follows:

$$\tilde{y} \cong \frac{1}{L}\sum_{i=I}^{R+L-1}F(M_i) \tag{13}$$

Here $F$ is the mathematical model that gives the output given the input, $\tilde{y}$ is the average prediction of the Bayesian rough set model, $R$ is the number of initial states that are discarded in the hope of reaching a stationary posterior distribution function described in equation 11 and $L$ is the number of retained states. In this paper, MCMC method is implemented by sampling a stochastic process consisting of random variables $\{g_1, g_2, ..., g_n\}$ through introducing random changes to granule vector $\{g\}$ and either accepting or rejecting the state according to Metropolis et al. algorithm given the differences in posterior probabilities between two states that are in transition (Metropolis et al., 1953). This algorithm ensures that states with high probability form the majority of the Markov chain and is mathematically represented as:

If $P(M_{n+1} | D) > P(M_n | D)$ then accept $M_{n+1}$, (14)

else accept if $P(M_{n+1} | D) / P(M_n | D) > \xi$ where $\xi \in [0,1]$ (15)

else reject and randomly generate another model $M_{n+1}$.

**Experimental Investigation: Modelling of HIV**

The proposed method is applied to create a model that uses demographic characteristics to estimate the risk of HIV. In the last 20 years, over 60 million people have been infected with HIV (Human immunodeficiency virus), and of those cases, 95% are in developing countries (Lasry et al, 2007). HIV has been identified as the cause of AIDS. Early studies on HIV/AIDS focused on the individual characteristics and behaviors in determining HIV risk and Fee and Krieger (1993) refer to this as biomedical individualism. But it has been determined that the study of the distribution of health outcomes and their social determinants is of more importance and this is referred to as social epidemiology (Poundstone et. al., 2004). This study uses individual characteristics as well as social and demographic factors in determining the risk of HIV using rough set models formulated using Bayesian approach and trained using Monte Carlo method. Previously, computational intelligence techniques have been used extensively to analyze HIV and Leke et al (2006, 2006, 2007) used autoencoder network classifiers, inverse neural networks, as well as conventional feed-forward neural networks to estimate HIV risk from demographic factors. Although good accuracy was achieved when using the autoencoder method, it is disadvantageous due to its "black box" nature which is that it is not transparent. To improve transparency Bayesian rough set theory (RST) is proposed to forecast and interpret the causal effects of HIV. Rough set have been used in various biomedical and engineering applications (Ohrn, 1999; Pe-a et. al, 1999; Tay and Shen, 2003; Golan and Ziarko, 1995). But in most applications, RST is used primarily for prediction and this paper proposes Bayesian rough set models for HIV prediction. Rowland et al (1998) compared the use of RST and neural networks for the prediction of ambulation spinal cord injury, and although the neural network method produced more accurate results, its "black box" nature makes it impractical for the use of rule extraction problems. Poundstone et al (2004) related demographic properties to the spread of HIV. In their work they justified the use of demographic properties to create a model to predict HIV from a given database, as is done in this study. In order to achieve good accuracy, the rough set partitions or discretisation process needs to be well chosen and this is done by sampling through the granulization space and accepting the samples with high posterior probability using Metropolis et. al. algorithms (1953). The data set used in this paper was obtained from the South African antenatal sero-prevalence survey of 2001 (Department of Health, 2001). The data was obtained through questionnaires completed by pregnant women attending selected public clinics and was conducted concurrently across all nine provinces in South Africa. The six demographic variables considered are: race, age of mother, education, gravidity, parity and, age of father, with the outcome or decision being either HIV positive or negative. The HIV status is the decision represented in binary form as either a 0 or 1, with a 0 representing HIV negative and a 1 representing HIV positive. The input data was discretised into four partitions. This number was chosen as it gave a good balance between computational efficiency and accuracy. The parents' ages are given and discretised accordingly, education is given as an integer, where 13 is the highest level of education, indicating tertiary education. Gravidity is defined as the number of times that a woman has been pregnant, whereas parity is defined as the number of times that she has given birth. It must be noted that multiple births during a pregnancy are indicated with a parity of one. Gravidity and parity also provide a good indication of the reproductive health of pregnant women in South Africa. The rough set models were trained by sampling in the input space and accepting or rejecting using Metropolis et. al. algorithm (1953). The sample input space and the results therefore are shown in Table 2.

Table 2: The input space partitioned as Low, Med, High and Very High

| LowA | MedA | MedB | HighB | MedC | ... | HighD | LowE | HighE | Accuracy | Number Rules |
|---|---|---|---|---|---|---|---|---|---|---|
| 6.14 | 27.03 | 5.86 | 9.31 | 1.63 | ... | 2.56 | 2.38 | 10.85 | 55.50 | 191.00 |
| 11.44 | 15.77 | 9.21 | 10.19 | 2.76 | ... | 5.71 | 0.59 | 32.67 | 59.87 | 299.00 |
| 8.56 | 24.08 | 7.01 | 8.10 | 4.62 | ... | 3.65 | 7.83 | 28.55 | 56.44 | 202.00 |
| 1.78 | 3.76 | 0.00 | 1.71 | 0.27 | ... | 3.39 | 6.54 | 19.84 | 60.77 | 130.00 |
| 4.12 | 6.33 | 1.25 | 6.86 | 1.77 | ... | 4.15 | 10.37 | 28.81 | 57.52 | 226.00 |
| 7.83 | 20.49 | 1.45 | 4.99 | 1.13 | ... | 3.36 | 5.00 | 23.70 | 62.54 | 283.00 |
| 2.68 | 25.31 | 4.98 | 6.24 | 0.32 | ... | 3.72 | 0.79 | 14.97 | 56.37 | 204.00 |

As with many surveys, there are incomplete entries and such cases are removed from the data set. The second irregularity was information that is false for example an instance where gravidity (number of pregnancies) was zero and parity (number of births) was at least one, which is impossible because for a woman to have given birth she must not have been pregnant. Such cases were removed from the data set. Only 12945 cases remained from a total of 13087. The input data was therefore the demographic characteristics explained earlier and the output were the plausibility of HIV with 1 representing 100% plausibility that a person is HIV positive and -1 indicating 100% plausibility of HIV negative. When training the rough set models using Markov Chain Monte Carlo, 500 samples were accepted and retained meaning that 500 sets of rules where each set contained 50 up to 550 numbers of rules with an average of 222 rules as can be seen in Figure 1. 500 samples were retained because the simulation had converged to a stationary distribution. This figure must be interpreted in the light of the fact on calculating the posterior probability we used the knowledge that fewer rules are more desirable than many. Therefore, the Bayesian rough set framework is able to select the number of rules in addition to the partition sizes.

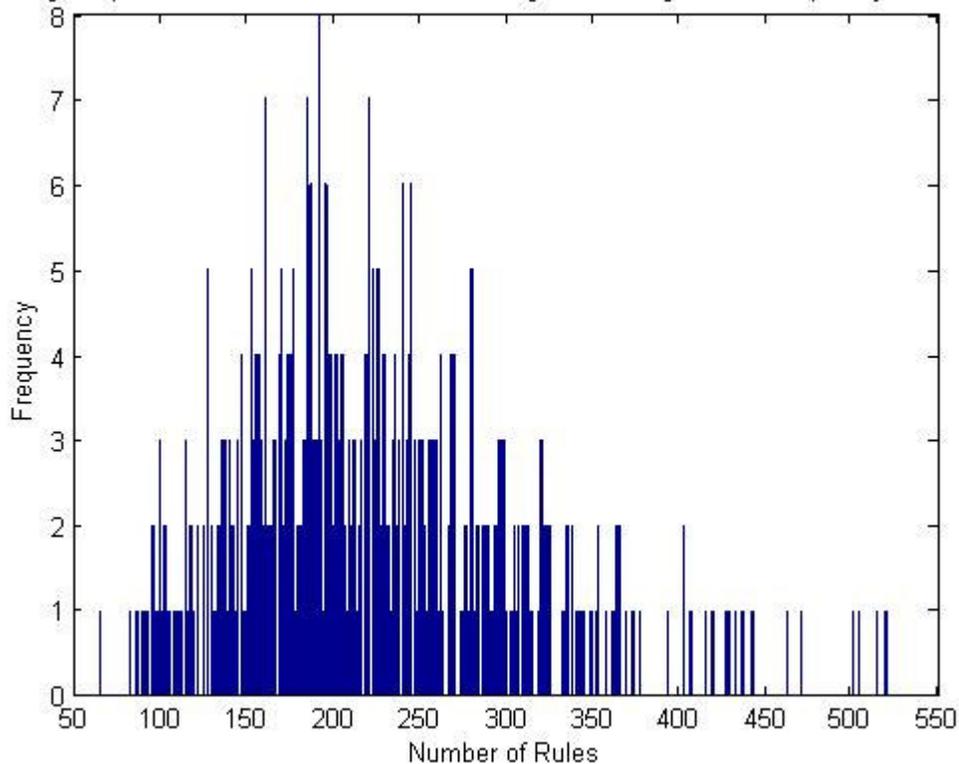

Figure 1: The distribution of number of rules

The average accuracy achieved was 58% and while the accuracy attained varied from 50% up to 66% as can be seen from Figure 2.

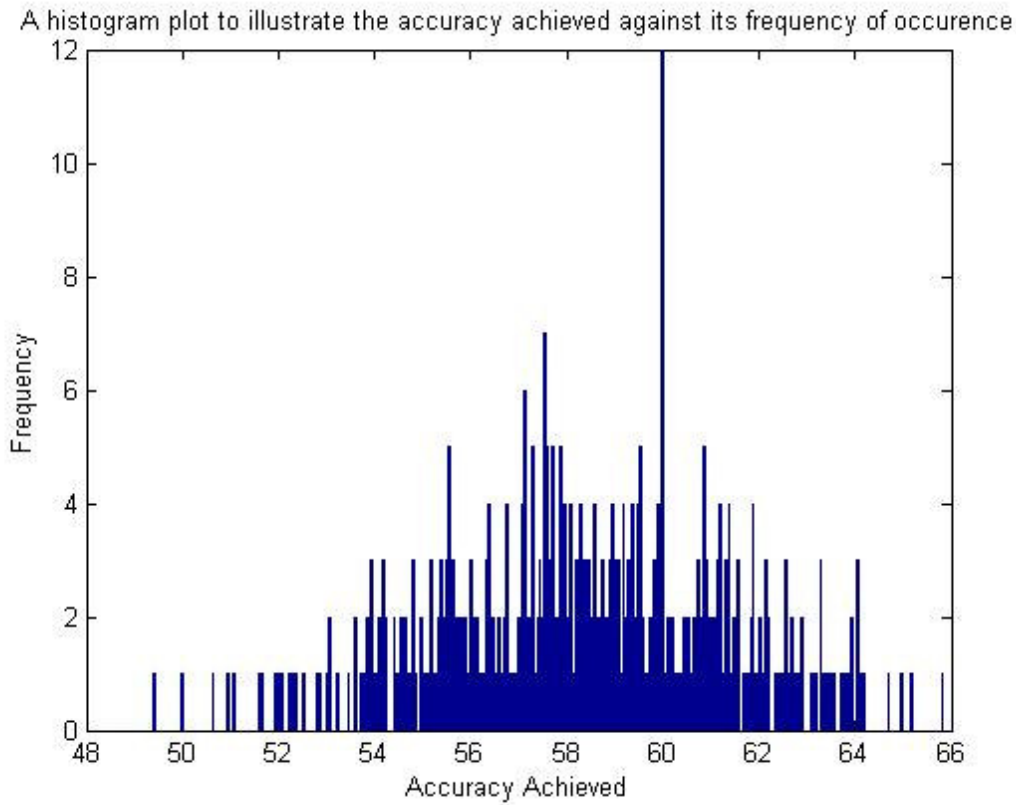

Figure 2: A histogram showing the accuracy achieved versus the frequency of occurrence

Traditional rough set give the plausibility such as 0.46 (46%) plausibility that a person is HIV positive. Bayesian rough set approach allows us to determine how confident we are of that plausibility. For example, Figure 3 shows that the average plausibility achieved is -0.48 indicating 48% plausibility that a person is HIV negative. In addition as can be seen in Figure 3, we can determine the probability distribution of that plausibility. This in essence indicate that the Bayesian rough set models allow us to interpret the predictions of rough set models in probability terms as can be viewed from a probability distribution in Figure 3.

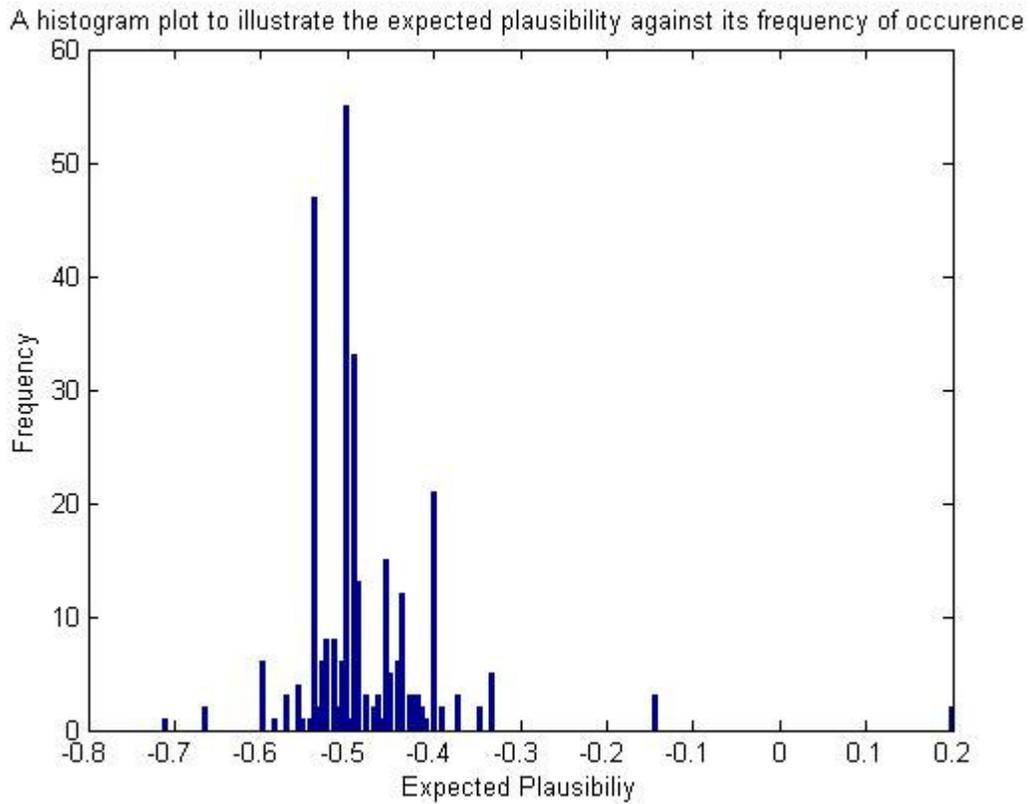

Figure 3: Plausability distribution of an HIV negative outcome.

*Rule Extraction*

Once Bayesian RST was applied to the HIV data, unique distinguishable cases and indiscernible cases were extracted. From the data set of 12945 cases, the data is only a representative of 452 cases out of the possible 4096 unique combinations. The lower approximation cases are rules that always hold, or are definite cases while the upper approximation can only be stated with certain plausibility. Examples of both cases that were extracted from the approach in this paper are stated below:

*Lower Approximation Rules*

1. If Race = African and Mothers Age = 23 and Education = 4 and Gravidity = 2 and Parity = 1 and Fathers Age = 20 Then HIV = Most Probably Positive

2. If Race = Asian and Mothers Age = 30 and Education = 13 and Gravidity = 1 and Parity = 1 and Fathers Age = 33 Then HIV = Most Probably Negative

*Upper Approximation Rules*

1. If Race = Coloured and Mothers Age = 33 and Education = 7 and Gravidity = 1 and Parity = 1 and Fathers Age = 30 Then HIV = Positive with plausibility = 0.33333

2. If Race = White and Mothers Age = 20 and Education = 5 and Gravidity = 2 and Parity = 1 and Fathers Age = 20 Then HIV = Positive with plausibility = 0.06666

**Conclusion**

Rough set were formulated using Bayesian framework. They were then trained using Markov Chain Monte Carlo method. The Bayesian framework is found to offer probabilistic interpretations to rough set. A balance between transparency of the rough set model and the accuracy of HIV estimation is achieved with a great deal of computational effort.

**References**

1. Bishop, C.M., 2006. Pattern recognition and machine intelligence. Springer, Berlin, Germany.